\documentclass[runningheads]{llncs}

\usepackage[T1]{fontenc}
\usepackage{longtable}
\newlength\savewidth\newcommand\shline{\noalign{\global\savewidth\arrayrulewidth
  \global\arrayrulewidth 1.5pt}\hline\noalign{\global\arrayrulewidth\savewidth}}
\usepackage{threeparttable}
\usepackage{graphicx}
\usepackage{framed,multirow}
\usepackage{amsmath}
\usepackage{booktabs}
\usepackage{algorithm}
\usepackage{makecell}
\usepackage[colorlinks=true,
    linkcolor=red,
    urlcolor=blue,citecolor=blue]{hyperref}
\usepackage{xspace}
\newcommand{\methodname}{\texttt{BerDiff}\xspace}
\usepackage{amsmath,amsfonts}
\usepackage{amssymb,amsopn}
\usepackage{algorithm}
\usepackage{algpseudocode}
\usepackage{bm}
\usepackage{bbm}

\newcommand{\std}[1]{\tiny #1}
\newcommand{\ie}{\textit{i}.\textit{e}.\xspace}
\newcommand{\eg}{\textit{e}.\textit{g}.\xspace}

\newcommand{\eat}[1]{}

\newcommand{\tabref}[1]{Table~\ref{#1}}
\newcommand{\figref}[1]{Fig.~\ref{#1}}

\newcommand{\equref}[1]{Eq.~\eqref{#1}}
\newcommand{\secref}[1]{Sec.~\ref{#1}}

\newcommand{\tabincell}[2]{\begin{tabular}{@{}#1@{}}#2\end{tabular}}

\newcommand{\loss}[1]{\mathcal{L}_\text{#1}}
\newcommand{\lamda}[1]{\lambda_\text{#1}}   

\newcommand{\best}[1]{\textbf{#1}}
\newcommand{\suboptimal}[1]{\underline{#1}}

\newcommand{\vct}[1]{\boldsymbol{#1}} 
\newcommand{\mat}[1]{\boldsymbol{#1}} 
\newcommand{\img}[1]{\boldsymbol{#1}} 

\begin{document}
\title{BerDiff: Conditional Bernoulli Diffusion Model for Medical Image Segmentation}
\titlerunning{Bernoulli Diffusion Model for Medical Image Segmentation}
\author{Tao Chen, Chenhui Wang, Hongming Shan}
\institute{Institute of Science and Technology for Brain-inspired Intelligence\\
Fudan University}
\maketitle
\begin{abstract}
Medical image segmentation is a challenging task with inherent ambiguity and high uncertainty, attributed to factors such as unclear tumor boundaries and multiple plausible annotations. The \emph{accuracy} and \emph{diversity} of segmentation masks are both crucial for providing valuable references to radiologists in clinical practice. While existing diffusion models have shown strong capacities in various visual generation tasks, it is still challenging to deal with discrete masks in segmentation. To achieve accurate and diverse medical image segmentation masks, we propose a novel conditional \textbf{Ber}noulli \textbf{Diff}usion model for medical image segmentation (\methodname). Instead of using the Gaussian noise, we first propose to use the Bernoulli noise as the diffusion kernel to enhance the capacity of the diffusion model for binary segmentation tasks, resulting in more accurate segmentation masks. Second, by leveraging the stochastic nature of the diffusion model, our \methodname randomly samples the initial Bernoulli noise and intermediate latent variables multiple times to produce a range of diverse segmentation masks, which can highlight salient regions of interest that can serve as valuable references for radiologists. In addition, our \methodname can efficiently sample sub-sequences from the overall trajectory of the reverse diffusion, thereby speeding up the segmentation process. Extensive experimental results on two medical image segmentation datasets with different modalities demonstrate that our \methodname outperforms other recently published state-of-the-art methods. Our results suggest diffusion models could serve as a strong backbone for medical image segmentation.
\end{abstract}
\section{Introduction}
Medical image segmentation plays a crucial role in enabling better diagnosis, surgical planning, and image-guided surgery~\cite{haque2020deep}. The inherent ambiguity and high uncertainty of medical images pose significant challenges~\cite{baumgartner2019phiseg} for accurate segmentation, attributed to factors such as unclear tumor boundaries in brain Magnetic resonance imaging (MRI) images and multiple plausible annotations in lung nodule Computed Tomography (CT) images. Existing medical image segmentation works typically provide a single, deterministic, most likely hypothesis mask, which may lead to misdiagnosis or sub-optimal treatment. Therefore, providing \emph{accurate} and \emph{diverse} segmentation masks as valuable references~\cite{lenchik2019automated} for radiologists is crucial in clinical practice.

Recently, diffusion models~\cite{ho2020denoising} have shown strong capacities in various visual generation tasks~\cite{ramesh2022hierarchical,saharia2022photorealistic}. However, how to better integrate with discrete segmentation tasks needs further consideration. Although many researches~\cite{amit2021segdiff,wu2022medsegdiff} have combined diffusion model with segmentation tasks and made some modifications, these methods do not take full account of the discrete characteristic of segmentation task and still use Gaussian noise as their diffusion kernel.
To achieve accurate and diverse segmentation, we propose a novel Conditional \textbf{Ber}noulli \textbf{Diff}usion model for medical image segmentation (\methodname). Instead of using the Gaussian noise, we first propose to use the Bernoulli noise as the diffusion kernel to enhance the capacity of the diffusion model for segmentation, resulting in more accurate segmentation masks.  Moreover, by leveraging the stochastic nature of the diffusion model, our \methodname randomly samples the initial  Bernoulli noise and intermediate latent variables multiple times to produce a range of diverse segmentation masks, which can highlight salient regions of interest (ROI) that can serve as a valuable reference for radiologists. In addition, our \methodname can efficiently sample sub-sequences from the overall trajectory of the reverse diffusion based on the rationale behind the Denoising Diffusion Implicit Models (DDIM)~\cite{wolleb2021diffusion}, thereby speeding up the segmentation process.

The contributions of this work are summarized as follows. 1) Instead of using the Gaussian noise, we propose a novel conditional diffusion model based on the Bernoulli noise for discrete binary segmentation tasks, achieving accurate and diverse medical image segmentation masks. 2) Our \methodname can efficiently sample sub-sequences from the overall trajectory of the reverse diffusion, thereby speeding up the segmentation process. 3) Experimental results on two medical images, CT and MRI, specifically the LIDC-IDRI and BRATS~2021 datasets, demonstrate that our \methodname outperforms other state-of-the-art methods.

\section{Methodology}
\begin{figure}[ht]
\centering
  \includegraphics[width=1\linewidth]{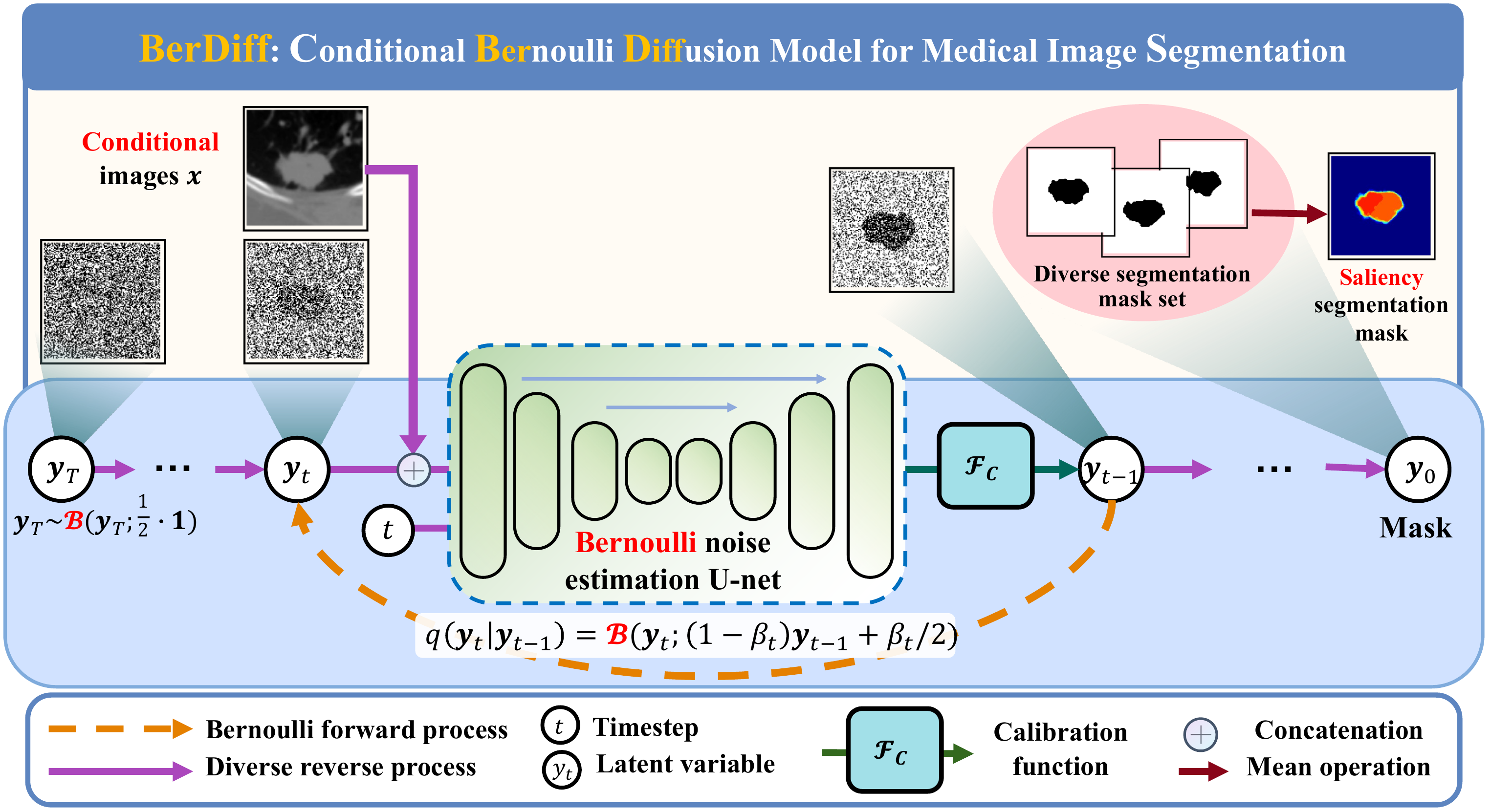}
  \caption{Illustration of Bernoulli forward and diverse reverse processes of our \methodname.}
  \label{fig:overviews}
\end{figure}

In this section, we first describe the problem definitions, then demonstrate the Bernoulli forward and diverse reverse processes of our \methodname, as shown in \figref{fig:overviews}. Finally, we provide an overview of the training and sampling procedures.

\subsection{Problem definition}

Let us assume that $\img{x} \in{\mathbb{R}^{H\!\times\! W\!\times\!C}}$ denotes the input medical image with a spatial resolution of $H\!\times\!W$ and $C$ number of channels. The ground-truth mask is represented as $\img{y}_{0}\!\in\!{{\{0,1\}}^{H\times W}}$, where  $0$ represents background while $1$ ROI. Inspired by diffusion-based models such as denoising diffusion probabilistic model (DDPM) and DDIM, we propose a novel conditional Bernoulli diffusion model, which can be represented as $p_{\theta}({\mat{y}_{0}}|\mat{x}):=\int p_{\theta}(\mat{y}_{0:T}|\mat{x})\mathrm{d}\mat{y}_{1:T}$, where $\mat{y}_{1},\ldots,\mat{y}_{T}$ are latent variables of the same size as the mask $\mat{y}_{0}$. 
For medical binary segmentation tasks, the diverse reverse process of our \methodname starts from the initial Bernoulli noise $\img{y}_{T} \sim \mathcal{B}(\img{y}_{T};\frac{1}{2}\!\cdot\! {\mathbf{1}})$ and progresses through intermediate latent variables constrained by the input medical image $\vct{x}$ to produce segmentation masks, where $\mathbf{1}$ denotes an all-ones matrix of the size $H\!\times\!W$.

\subsection{Bernoulli forward process}
In previous generation-related diffusion models, Gaussian noise is progressively added with increasing timestep $t$. However, for segmentation tasks, the ground-truth masks are represented by discrete values. To address this, our \methodname gradually adds more Bernoulli noise using a noise schedule $\beta_{1},\ldots, \beta_{T}$, as shown in \figref{fig:overviews}. The Bernoulli forward process $q(\mat{y}_{1:T}|\mat{y}_{0})$ of our \methodname is a Markov chain, which can be represented as:
\begin{align}
\label{eqa:forward_1t}
    q\left(\img{y}_{1: T} \mid \img{y}_{0}\right):=&\prod\nolimits_{t=1}^{T}q\left(\img{y}_{t} \mid \img{y}_{t-1}\right), \\ 
    \label{eqa:forward_tt-1}
    q\left(\img{y}_{t} \mid \img{y}_{t-1}\right):=&\mathcal{B}(\img{y}_{t} ; (1-\beta_{t}) \img{y}_{t-1}+\beta_{t}/2),
\end{align}
where $\mathcal{B}$ denotes the  Bernoulli distribution with the probability parameters $(1-\beta_{t}) \img{y}_{t-1}+\beta_{t}/2$.
Using the notation $\alpha_{t}=1-\beta_{t}$ and $\bar{\alpha}_{t}= {\textstyle \prod_{\tau=1}^{t}} {\alpha}_{\tau}$, we can efficiently sample $\img{y}_{t}$ at an arbitrary timestep $t$ as follows: 
\begin{align}
\label{eqa:forward_arbitrary}
q\left(\img{y}_{t} \mid \img{y}_{0}\right)=\mathcal{B}(\img{y}_{t}; \bar{\alpha}_{t} \img{y}_{0}+(1-\bar{\alpha}_{t}) / 2)).
\end{align}

\begin{figure}[htbp]
\centering
\begin{minipage}{0.49\linewidth}
\begin{algorithm}[H]
\caption{Training}\label{alg:training}
\begin{algorithmic}
\Repeat
\State $(\img{x}, \vct{y}_{0}) \sim q\left(\img{x}, \vct{y}_{0}\right)$
\State $t \sim \mathrm{Uniform}(\{1,\dots,T\})$
\State $\vct{\epsilon} \sim \mathcal{B}(\vct{\epsilon}; (1-\bar{\alpha}_{t})/2)$
\State $\vct{y}_{t} = \vct{y}_{0} \oplus \vct{\epsilon}$
\State \textbf{Calculate \equref{eqa:posterior}} 
\State \textbf{Estimate} $\hat{\vct{\epsilon}}(\vct{y}_{t}, {t}, \img{x})$
\State \textbf{Calculate \equref{eqa:predicted_posterior}}

\State \textbf{Take} gradient descent on $\nabla_{\theta}(\loss{Total})$
\Until{converged}
\end{algorithmic}
\end{algorithm}
\end{minipage}
\hfil
\begin{minipage}{0.5\linewidth}
\begin{algorithm}[H]
\caption{Sampling}\label{alg:sampling}
\begin{algorithmic}
\State $\vct{y}_{T} \sim \mathcal{B}(\img{y}_{T};\frac{1}{2}\cdot{\mathbf{1}})$
\For{$t = T$ to $1$}
\State $\hat{\img{\mu}}(\img{y}_{t}, t, \img{x}) =
    \mathcal{F}_{C}(\img{y}_{t},\hat{\img{\epsilon}}(\img{y}_{t}, t, \img{x}))$

\State \textbf{For DDPM}:
\State $\vct{y}_{t-1}\!\sim\!\mathcal{B}(\vct{y}_{t-1}; \hat{\vct{\mu}}(\vct{y}_{t}, t, \img{x}))$
\State \textbf{For DDIM}:
\State $\vct{y}_{t-1}\!\sim\! \mathcal{B}(\vct{y}_{t-1}; \sigma_{t}\vct{y}_{t}+(\bar{\alpha}_{t-1}\!-\!\sigma_{t}\bar{\alpha}_{t})|\vct{y}_{t}\!-\!\hat{\vct{\epsilon}}(\vct{y}_{t}, t, \img{x})|+((1-\bar{\alpha}_{t-1})-(1-\bar{\alpha}_{t})\sigma_{t})/2)$
\EndFor
\State \Return{$\vct{y}_{0}$}
\end{algorithmic}
\end{algorithm}
\end{minipage}
\end{figure}

To ensure that the objective function described in \secref{sec:detailed} is tractable and easy to compute, we use the sampled Bernoulli noise $\img{\epsilon}\!\sim\!\mathcal{B}(\img{\epsilon}; \frac{1-\bar{\alpha}_{t}}{2} \!\cdot\! {\mathbf{1}})$ to reparameterize $\img{y}_{t}$ of \equref{eqa:forward_arbitrary} as $\img{y}_{0}\oplus\img{\epsilon}$, where $\oplus$ denotes the logical operation of ``exclusive or (XOR)''. Additionally, let $\odot$ denote elementwise product, and $\operatorname{Norm}(\cdot)$ denote  normalizing the input data along the channel dimension and then returning the second channel. The concrete Bernoulli posterior can be represented as:
\begin{align}
\label{eqa:posterior}
q(\img{y}_{t-1}\mid\img{y}_{t}, \img{y}_{0})  &=\mathcal{B}(\img{y}_{t-1} ; \theta_{\text{post}}\left(\img{y}_{t}, \img{y}_{0}\right)).
\end{align}
where $\theta_{\text{post}}\left(\vct{y}_{t}, \vct{y}_{0}\right)=\operatorname{Norm}([\alpha_{t}[1-\img{y}_{t}, \img{y}_{t}]+\frac{1-\alpha_{t}}{2}] \odot [\bar{\alpha}_{t-1}[1-\img{y}_{0}, \img{y}_{0}]+\frac{1-\bar{\alpha}_{t-1}}{2}])$.

\subsection{Diverse reverse process}
The diverse reverse process $p_{\theta}(\img{y}_{0:T})$ can be also viewed as a Markov chain that starts from the Bernoulli noise $\img{y}_{T} \sim \mathcal{B}(\img{y}_{T};\frac{1}{2}\cdot{\mathbf{1}})$ and progresses through intermediate latent variables constrained by the input medical image $\vct{x}$ to produce diverse segmentation masks, as shown in \figref{fig:overviews}. The concrete diverse reverse process of our \methodname can be represented as:
\begin{align}
    p_{\theta}(\img{y}_{0: T}\mid \img{x})&:=p(\img{y}_{T}) \prod\nolimits_{t=1}^{T} p_{\theta}(\img{y}_{t-1} \mid \img{y}_{t}, \img{x}),\\
    \label{eqa:predicted_posterior}
    p_{\theta}(\img{y}_{t-1} \mid \img{y}_{t}, \img{x})&:=\mathcal{B}(\img{y}_{t-1}; \hat{\img{\mu}}(\img{y}_{t}, t, \img{x})).
\end{align}

Specifically, we utilize the estimated Bernoulli noise $\hat{\img{\epsilon}}(\img{y}_{t}, t, \img{x})$ of $\img{y}_{t}$ to parameterize $\hat{\img{\mu}}(\img{y}_{t}, t, \img{x})$ via a calibration function $\mathcal{F}_{C}$, as follows:
\begin{align}
    \hat{\img{\mu}}(\img{y}_{t}, t, \img{x}) =
    \mathcal{F}_{C}(\img{y}_{t},\hat{\img{\epsilon}}(\img{y}_{t}, t, \img{x}))=
    \theta_{\text {post}}(\img{y}_{t}, |\img{y}_{t}-\hat{\img{\epsilon}}(\img{y}_{t}, t, \img{x})|),
\end{align}
where $|\cdot|$ denotes the absolute value operation. 

\subsection{Detailed procedure}
\label{sec:detailed}
Here, we provide an overview of the training and sampling procedure in Algorithms~\ref{alg:training} and \ref{alg:sampling}. During the training phase, given an image and mask data pair $\{\img{x},\vct{y}_{0}\}$, we sample a random timestep $t$ from a uniform distribution $\{1,\dots,T\}$,  which is used to sample the Bernoulli noise $\img{\epsilon}$. 

We then use $\img{\epsilon}$ to sample $\vct{y}_{t}$ from $q(\vct{y}_t\!\mid\!\vct{y}_0)$, which allows us to obtain the Bernoulli posterior $q(\img{y}_{t-1}\!\mid\!\img{y}_{t}, \img{y}_{0})$. We pass the estimated Bernoulli noise $\img{\hat{\epsilon}}(\img{y}_{t}, t, \img{x})$ through the calibration function $\mathcal{F}_{C}$ to parameterize $p_{\theta}(\img{y}_{t-1}\!\mid\!\img{y}_{t}, \img{x})$. Based on the variational upper bound on the negative log-likelihood in previous diffusion models~\cite{austin2021structured}, we adopt Kullback-Leibler (KL) divergence and binary cross-entropy (BCE) loss to optimize our \methodname as follows:
 \begin{align}
     \label{eqa:loss_KL}
     \loss{KL} &= \mathbb{E}_{q(\img{x}, \img{y}_{0})}
\mathbb{E}_{q(\img{y}_{t} \mid \img{y}_{0})}[D_{\mathrm{KL}}[q(\img{y}_{t-1} \mid \img{y}_{t}, \img{y}_{0}) \| p_{\theta}(\img{y}_{t-1} \mid \img{y}_{t}, \mat{x})]],
 \\
    \label{eqa:loss_bce}
    \loss{BCE} &= -\mathbb{E}_{(\img{\epsilon}, \hat{\img{\epsilon}})} {\textstyle \sum_{i,j}^{H,W}} 
[{\epsilon}_{i,j}\log_{}\hat{\epsilon}_{i,j}  + (1-\epsilon_{i,j})\log_{}{(1-\hat{\epsilon}_{i,j})}].
\end{align}
Finally, the overall objective function is presented as:
\begin{align}
\label{equ:loss_final}
\loss{Total}=\loss{KL} + \lamda{BCE}\loss{BCE},
\end{align}
where  $\lamda{BCE}$ is set to $1$ in our experiments.

During the sampling phase, our \methodname first samples the initial latent variable $\img{y}_T$, followed by iterative calculation of the probability parameters of $\vct{y}_{t-1}$ for different $t$. In Algorithm~\ref{alg:sampling}, we present two different sampling strategies from DDPM and DDIM for the latent variable $\vct{y}_{t-1}$. Finally, our \methodname is capable of producing diverse segmentation masks. By taking the mean of these masks, we can further obtain a saliency segmentation mask to highlight salient ROI that can serve as a valuable reference for radiologists. Note that our \methodname proposes a novel parameterization technology, $\ie$ calibration function, to estimate the Bernoulli noise of $\img{y}_{t}$, which is different from previous discrete state diffusion-based models~\cite{austin2021structured,hoogeboom2021argmax,sohl2015deep}.

\section{Experiment}
\subsection{Experimental setup}
\noindent\textbf{Dataset and preprocessing}\quad The data used in this experiment are obtained from LIDC-IDRI~\cite{armato2011lung,clark2013cancer}  and BRATS~2021~\cite{Baid2021TheRB} datasets. LIDC-IDRI contains 1,018 lung CT scans with plausible segmentation masks annotated by four radiologists. We adopt a standard preprocessing pipeline for lung CT scans and the train-validation-test partition as in previous work~\cite{baumgartner2019phiseg,kohl2018probabilistic,selvan2020uncertainty}. BRATS~2021 consists of four different sequence (T1, T2, FlAIR, T1CE) MRI images for each patient. All 3D scans are sliced into axial slices and discarded the bottom 80 and top 26 slices. Note that we treat the original four types of brain tumors as one type following previous work~\cite{wolleb2021diffusion}, converting the multi-target segmentation problem into binary. Our training set includes 55,174 2D images scanned from 1,126 patients, and the test set comprises 3,991 2D images scanned from 125 patients. Finally, the images from LIDC-IDRI and BRAST~2021 are resized to  $128\times128$ and $224\times224$, respectively.

\begin{table}[!ht]
\centering
\caption{Ablation results of hyperparameters on LIDC-IDRI.}\label{table:ablation_hyper}
\begin{threeparttable}
\begin{tabular*}{1\linewidth}{@{\extracolsep{\fill}}llccccc}
\shline
\multirow{2}{*}{\textbf{Loss}}&\textbf{Estimation}&\multicolumn{4}{c}{\textbf{GED}}&\textbf{HM-IoU}\\
&\textbf{Target}&\textbf{16}&\textbf{8}&\textbf{4}&\textbf{1}&\textbf{16}\\
\cline{1-2}\cline{3-6}\cline{7-7}

$\loss{KL}$ & Bernoulli noise  & 0.332 & 0.365 & 0.430 & 0.825 & 0.517 \\
$\loss{BCE}$& Bernoulli noise  & \suboptimal{0.251} & \suboptimal{0.287} & \suboptimal{0.359} & \suboptimal{0.785} & \suboptimal{0.566} \\
$\loss{BCE}+\loss{KL}$ & Bernoulli noise  & $\best{0.249}$ & $\best{0.287}$ & \best{0.358} & \best{0.775} & $\best{0.575}$ \\
$\loss{BCE}+\loss{KL}$ & Ground-truth mask  & 0.277 & 0.317 & 0.396 & 0.866 & 0.509 \\ 

    \shline
     \end{tabular*}
     \end{threeparttable}
\end{table}
\begin{table}[!ht]
\centering
\caption{Ablation results of diffusion kernel on LIDC-IDRI}\label{table:ablation_kernel}
\begin{threeparttable}
\begin{tabular*}{1\linewidth}{@{\extracolsep{\fill}}clccccc}
\shline
\textbf{Training}&\textbf{Diffusion}&\multicolumn{4}{c}{\textbf{GED}}&\textbf{HM-IoU}\\
\textbf{Iteration}&\textbf{Kernel}&\textbf{16}&\textbf{8}&\textbf{4}&\textbf{1}&\textbf{16}\\
\cline{1-2}\cline{3-6}\cline{7-7}
\multirow{2}{*}{21,000}&Gaussian & 0.671 & 0.732 & 0.852 & 1.573 & 0.020 \\
&Bernoulli& \best{0.252} & \best{0.287} & \best{0.358} & \best{0.775} & \best{0.575}\\
   \hline
\multirow{2}{*}{86,500}  &Gaussian & 0.251 & 0.282 & 0.345 & \best{0.719} & 0.587 \\
   &Bernoulli & \best{0.238} & \best{0.271} & \best{0.340} & 0.748 & \best{0.596} \\
    \shline
     \end{tabular*}
     \end{threeparttable}
\end{table}

 \noindent\textbf{Implementation Details}\quad We implement all the methods with the PyTorch library and train the models on NVIDIA V100 GPUs. All the networks are trained using the AdamW~\cite{loshchilov2017decoupled} optimizer with a batch size of 32. The initial learning rate is set to $1\times 10^{-4}$ for BRATS~2021 and $5\times 10^{-5}$ for LIDC-IDRI. The Bernoulli noise estimation U-net network in \figref{fig:overviews} of our \methodname is the same as previous diffusion-based models~\cite{nichol2021improved}. We employ a linear noise schedule for $T=1000$ timesteps for all the diffusion models. And we use the sub-sequence sampling strategy of DDIM to accelerate the segmentation process. During mini-batch training of LIDC-IDRI, our \methodname learn diverse expertise by randomly sampling one from four annotated segmentation masks for each image. Three metrics are used for performance evaluation, including Generalized Energy Distance (GED), Hungarian-matched Intersection over Union (HM-IoU), and Dice coefficient. We compute GED using a varying number of segmentation samples (1, 4, 8, and 16), HM-IoU using 16 samples.

\subsection{Ablation study}
We start by conducting ablation experiments to demonstrate the effectiveness of different losses and estimation targets, as shown in \tabref{table:ablation_hyper}. All experiments are trained for 21,000 training iterations on LIDC-IDRI.  We first conduct the ablation study of different losses while estimating Bernoulli noise in the top three rows. We find that the combination of KL divergence and BCE loss can achieve the best performance. Then, we conduct an ablation study of selecting estimation target in the bottom two rows. We observe that estimating Bernoulli noise, instead of directly estimating the ground-truth mask, is more suitable for our binary segmentation task. All of these findings are consistent with previous works~\cite{austin2021structured,ho2020denoising}. Please refer to Appendix~\ref{appendix:strategy_timestep} for extra ablation studies on the sampling strategy and sampled timesteps.

\begin{figure}[ht]
\centering
\begin{minipage}[t]{0.52\linewidth}
\makeatletter\def\@captype{table}
\caption{Results on LIDC-IDRI.}
\label{tab:quantitative_CT}
\begin{threeparttable}
\begin{tabular*}{1\linewidth}{@{\extracolsep{\fill}}lcc}
\shline
\textbf{Methods} &  \tabincell{c}{\textbf{GED}\\\textbf{16}}
& \tabincell{c}{\textbf{HM-IoU}\\\textbf{16}} \\
\cline{1-3}
Prob.U-net~\cite{kohl2018probabilistic} &  0.320\std{$\pm$0.03}   & 0.500\std{$\pm$0.03}\\
Hprob.U-net~\cite{kohl2019hierarchical} &  0.270\std{$\pm$0.01}   & 0.530\std{$\pm$0.01}\\ 
CAR~\cite{kassapis2021calibrated}       &  0.264\std{$\pm$0.00}    &0.592\std{$\pm$0.01}\\
JPro.U-net~\cite{zhang2022probabilistic} &  0.260\std{$\pm$0.00}  &0.585\std{$\pm$0.00} \\
PixelSeg~\cite{zhang2022pixelseg}       &  0.260\std{$\pm$0.00} &0.587\std{$\pm$0.01} \\
SegDiff~\cite{amit2021segdiff} & \suboptimal{0.248\std{$\pm$0.01}}  & \suboptimal{0.585\std{$\pm$0.00}} \\
MedSegDiff~\cite{wu2022medsegdiff}& 0.420\std{$\pm$0.03}  &  0.413\std{$\pm$0.03} \\
BerDiff (\textbf{ours}) & \best{0.238\std{$\pm$0.01}}  &  \best{0.596\std{$\pm$0.00}}\\

\shline
\end{tabular*}
\end{threeparttable}
\end{minipage}
\begin{minipage}[t]{0.4\linewidth}
\makeatletter\def\@captype{table}
\caption{Results on BRATS~2021.}
\label{tab:quantitative_MRI}
\begin{threeparttable}
\begin{tabular*}{1\linewidth}{@{\extracolsep{\fill}}lc}
\shline
\multirow{2}{*}{\textbf{Methods}}&\multirow{2}{*}{\textbf{Dice}}\\
&\\
\cline{1-2}
nnU-net~\cite{isensee2021nnu} & 88.2 \\
TransU-net~\cite{chen2021transunet} & 88.6 \\
Swin UNETR~\cite{hatamizadeh2022swin} & 89.0 \\
U-net$^\sharp$& 89.2 \\
SegDiff~\cite{amit2021segdiff} & \suboptimal{89.3} \\
BerDiff (\textbf{ours}) & \best{89.7} \\
\shline
\end{tabular*}
\end{threeparttable}
\begin{tablenotes}[flushleft]\tiny
\item $\sharp$ The U-net has the same architecture as the noise estimation network in our \methodname and previous diffusion-based models.
\end{tablenotes}
\end{minipage}
\end{figure}

\begin{figure}[ht]
\centering
  \includegraphics[width=1\linewidth]{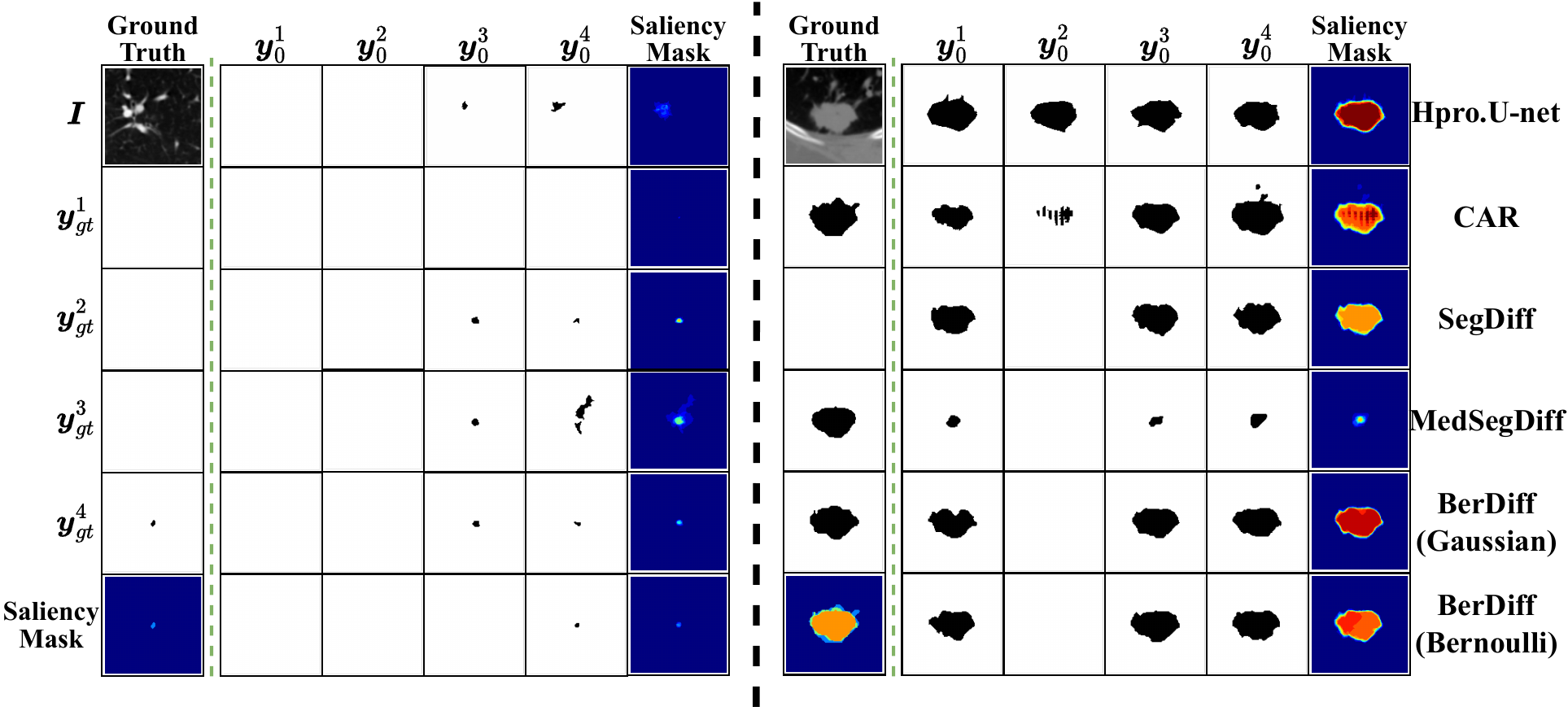}
  \caption{\textbf{Diverse segmentation masks and the corresponding saliency mask of two lung nodules randomly selected in LIDC-IDRI.}
   $\img{x}^{i}_{0}$ and $\img{x}^{i}_{\text{gt}}$ refer to the $i$-th generated and ground-truth segmentation masks, respectively. Saliency Mask is the mean of diverse segmentation masks.
}
  \label{fig:visualization_LIDC}
\end{figure}

Here, we conduct ablation experiments on our \methodname with Gaussian or Bernoulli noise, and the results are shown in \tabref{table:ablation_kernel}. For discrete segmentation tasks, we find that using Bernoulli noise can produce favorable results when training iterations are limited (\eg 21,000 iterations), and even outperform using Gaussian noise when training iterations are sufficient (\eg 86,500 iterations). We also provide a more detailed performance comparison between Bernoulli- and Gaussian-based diffusion models over training iterations in Appendix~\ref{appendix:performance_curves}. In addition, we present a toy example to demonstrate the superiority of Bernoulli diffusion over Gaussian diffusion in Appendix~\ref{appendix:toy_example}. 

\subsection{Comparison to other state-of-the-art methods}

\begin{figure}[h]
\centering
  \includegraphics[width=0.9\linewidth]{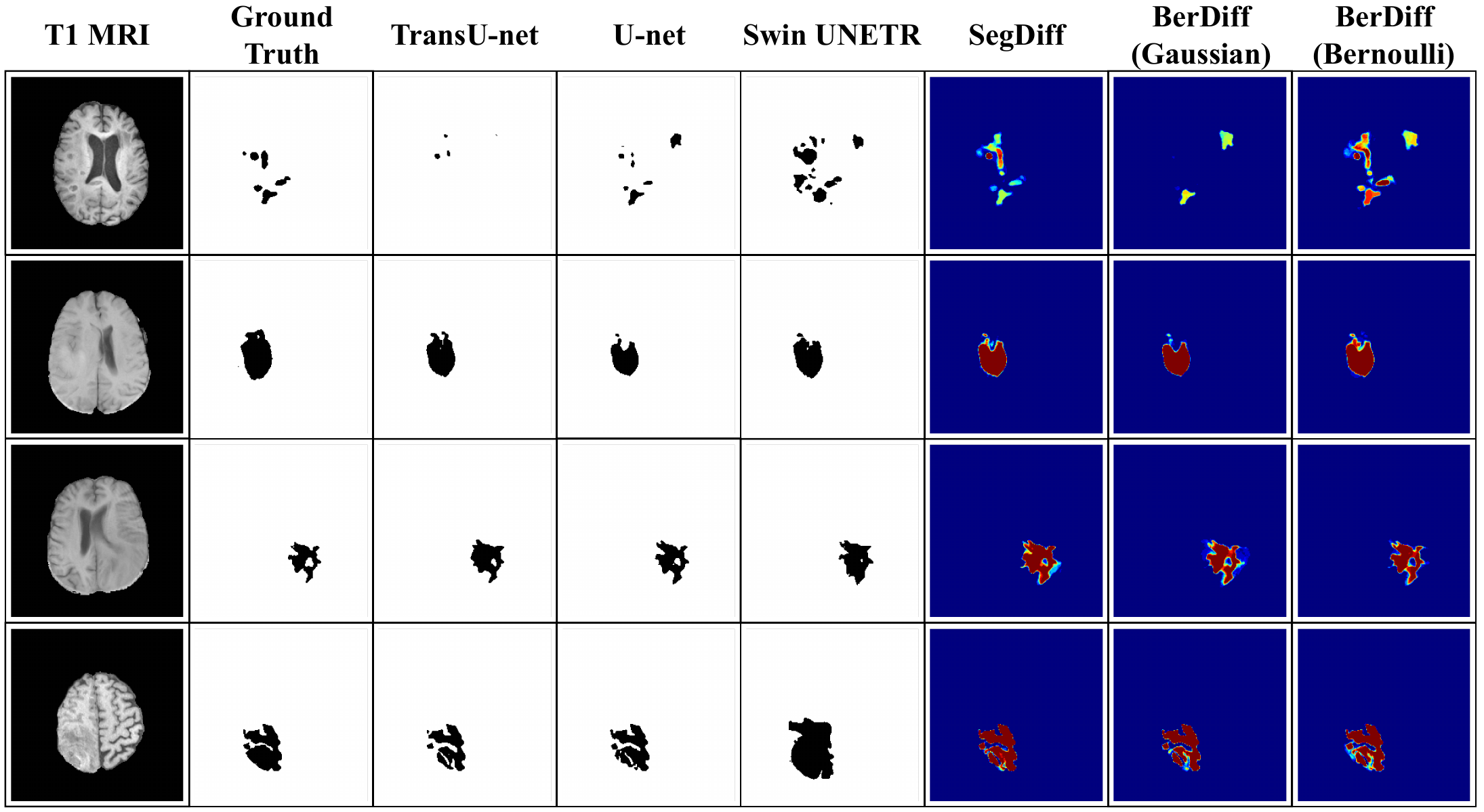}
  \caption{\textbf{Segmentation masks of four MRI images randomly selected in BRATS~2021.} The segmentation masks of diffusion-based models (SegDiff and ours) presented here are saliency segmentation masks.}
  \label{fig:visualization_BRATS}
\end{figure}
\noindent\textbf{Results on LIDC-IDRI}\quad Here, we present the quantitative comparison results of LIDC-IDRI in \tabref{tab:quantitative_CT}, and find that our \methodname perform well for discrete segmentation tasks. Probabilistic U-net (Prob.U-net), Hierarchical Prob.U-net (Hprob.U-net), and Joint Prob.U-net (JPro.U-net) use conditional variational autoencoder (cVAE) to accomplish segmentation tasks. Calibrated Adversarial Refinement (CAR) employs generative adversarial networks (GAN) to refine segmentation. PixelSeg is based on autoregressive models, while SegDiff and MedSegDiff are diffusion-based models. We have the following two observations: 1) diffusion-based methods have demonstrated significant superiority over traditional approaches based on VAE, GAN, and autoregression models for discrete segmentation tasks; and 2) our \methodname has outperformed other diffusion-based models that use Gaussian noise as the diffusion kernel. At the same time, we present comparison segmentation results in \figref{fig:visualization_LIDC}. Compared to other models, our \methodname can effectively learn diverse expertise, resulting in more diverse and accurate segmentation masks. Especially for small nodules that can create ambiguity, such as the lung nodule on the left, our \methodname approach produces segmentation masks that are more in line with the ground-truth masks.

\noindent\textbf{Results on BRATS~2021}\quad 
Here, we present the quantitative and qualitative comparison results of BRATS~2021 in \tabref{tab:quantitative_MRI} and \figref{fig:visualization_BRATS}, respectively. We conducted a comparative analysis of our \methodname with other models such as nnU-net, transformer-based models like TransU-net and Swin UNETR, as well as diffusion-based methods like SegDiff. First, we find that diffusion-based methods have shown superior performance compared to traditional U-net and transformer-based approaches. Besides, the high performance achieved by U-net, which shares the same architecture as our noise estimation network, highlights the effectiveness of the backbone design in diffusion-based models. Moreover, our proposed \methodname surpasses other diffusion-based models that use Gaussian noise as the diffusion kernel. Finally, from \figref{fig:visualization_BRATS}, we find that our \methodname segments more accurately on parts that are difficult to recognize by the human eye, such as the tumor in the $3$rd row. At the same time, we can also generate diverse plausible segmentation masks to produce a saliency segmentation mask. We note that some of these masks may be false positives as shown in the $1$st row, but they can be filtered out due to low saliency. Please refer to Appendix~\ref{appendix:more_example} for more examples of diverse segmentation masks generated by our \methodname.
\section{Conclusion}
We first propose to use the Bernoulli noise as the diffusion kernel to enhance the capacity of the diffusion model for binary segmentation tasks, achieving accurate and diverse medical image segmentation results. Our \methodname only focuses on binary segmentation tasks and takes much time during the iterative sampling process as other diffusion-based models; \eg our \methodname takes 0.4s to segment one medical image, which is ten times of traditional U-net. In the future, we will extend our \methodname to the multi-target segmentation problem and implement additional strategies for speeding up the segmentation process.

\newpage 
\section*{Appendix}
\appendix

\setcounter{figure}{0}
\setcounter{table}{0}
\setcounter{equation}{0}
\setcounter{algorithm}{0}
\setcounter{section}{0}
\renewcommand{\thefigure}{A\arabic{figure}}
\renewcommand{\thetable}{A\arabic{table}}

\section{Ablation study on sampling strategy and timestep}
\label{appendix:strategy_timestep}

Our \methodname is compatible with various sampling strategies, and here, we compare the performance of \methodname using DDPM's and DDIM's sampling strategies. The concrete sampling algorithms can be found in Algorithm~\ref{alg:sampling}. Our results  in \tabref{table:appendix_ablation_hyper} indicate that for binary segmentation tasks, \methodname using DDIM's sampling strategy achieves better performance compared to using DDPM's. Furthermore, to attain satisfactory performance with limited computational resources, we uniformly sample 10 timesteps from the complete trajectory in all other experiments.

\begin{table}[h]
\centering
\caption{\textbf{Ablation results of sampling strategy and timestep on LIDC-IDRI.} The model utilized in this study was trained for 21,000 training iterations.
}\label{table:appendix_ablation_hyper}
\begin{threeparttable}
\begin{tabular*}{1\linewidth}{@{\extracolsep{\fill}}lcccccc}
\shline
\multirow{2}{*}{\textbf{Configuration}}&\textbf{Sampled}&\multicolumn{4}{c}{\textbf{GED}}&\textbf{HM-IoU}\\
&\textbf{Timestep}&\textbf{16}&\textbf{8}&\textbf{4}&\textbf{1}&\textbf{16}\\
\cline{1-2}\cline{3-6}\cline{7-7}
\multirow{4}{*}{\makecell{\methodname\\ +\\ DDPM's sampling strategy}}
 & 2  & 0.441 & 0.483 & 0.568 & 1.076 & 0.303 \\
  & 10  & 0.266 & 0.302 & 0.377 & \best{0.824} & 0.533 \\
  & 100  & \suboptimal{0.258} & \suboptimal{0.296} & \suboptimal{0.372} & \suboptimal{0.829} & \suboptimal{0.539} \\
  & 1000  & \best{0.254} & \best{0.293} & \best{0.369} & 0.832 & \best{0.539} \\
\cline{1-2}\cline{3-6}\cline{7-7}
\multirow{4}{*}{\makecell{\methodname\\ + \\DDIM's sampling strategy}}
 & 2  & 0.432 & 0.481 & 0.579 & 1.167 & 0.341 \\ 
  & 10  & 0.252 & 0.287 & 0.358 & 0.775 & 0.575 \\ 
   & 100  & \suboptimal{0.250} & \suboptimal{0.284} & \suboptimal{0.351} & \suboptimal{0.759} & \suboptimal{0.582} \\ 
 & 1000 & \best{0.247} & \best{0.280} & \best{0.348} & \best{0.758} & \best{0.585} \\ 
\shline
\end{tabular*}
\end{threeparttable}
\end{table}
\begin{figure}[htbp]
\centering
  \includegraphics[width=1\linewidth]{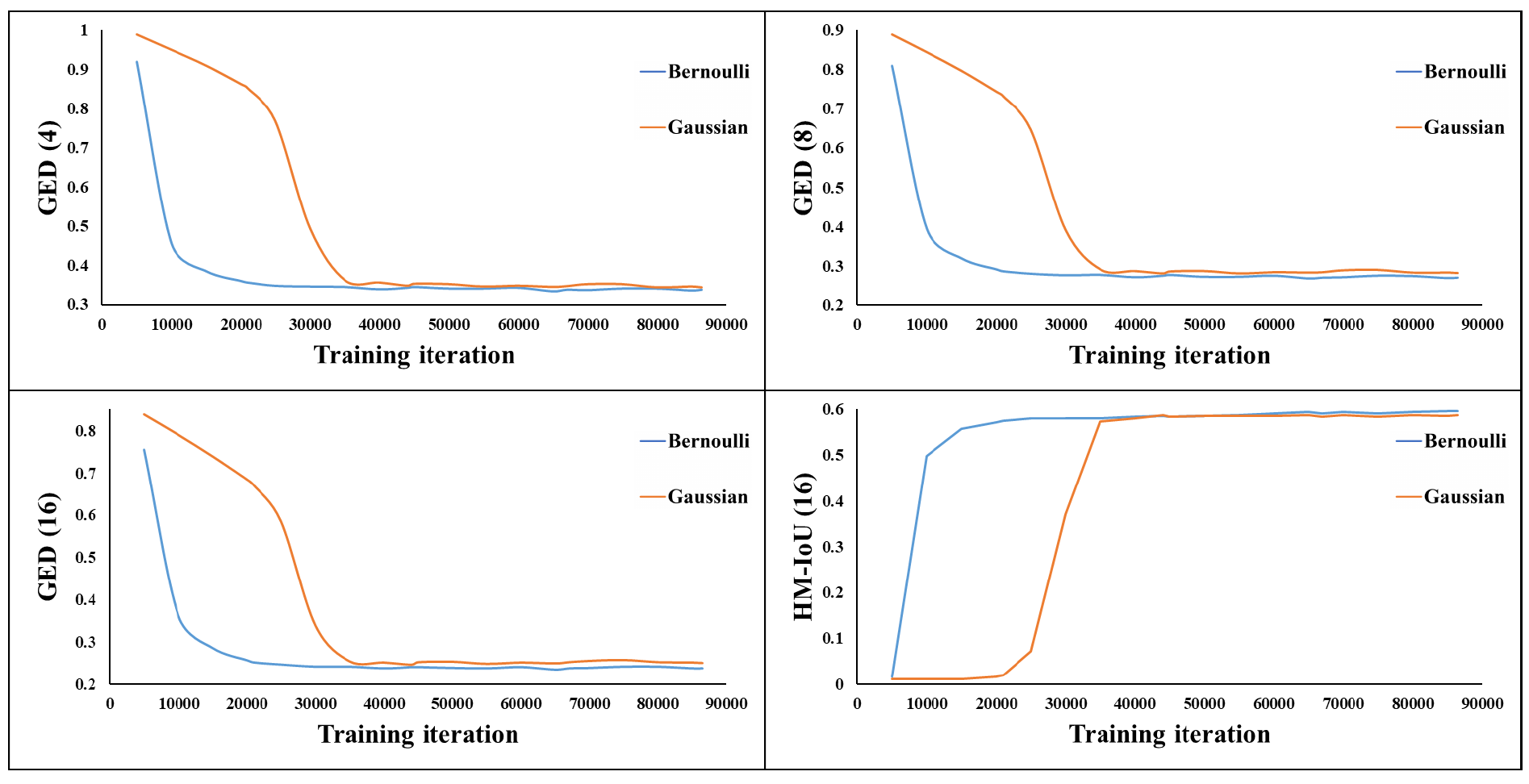}
  \caption{\textbf{Performance curves over training iterations for Gaussian- and Bernoulli-based diffusion models on LIDC-IDRI.} }
  \label{fig:appendix_fig_p_i_curve_supp}
\end{figure}
\begin{figure}[htbp]
\centering
  \includegraphics[width=1\linewidth]{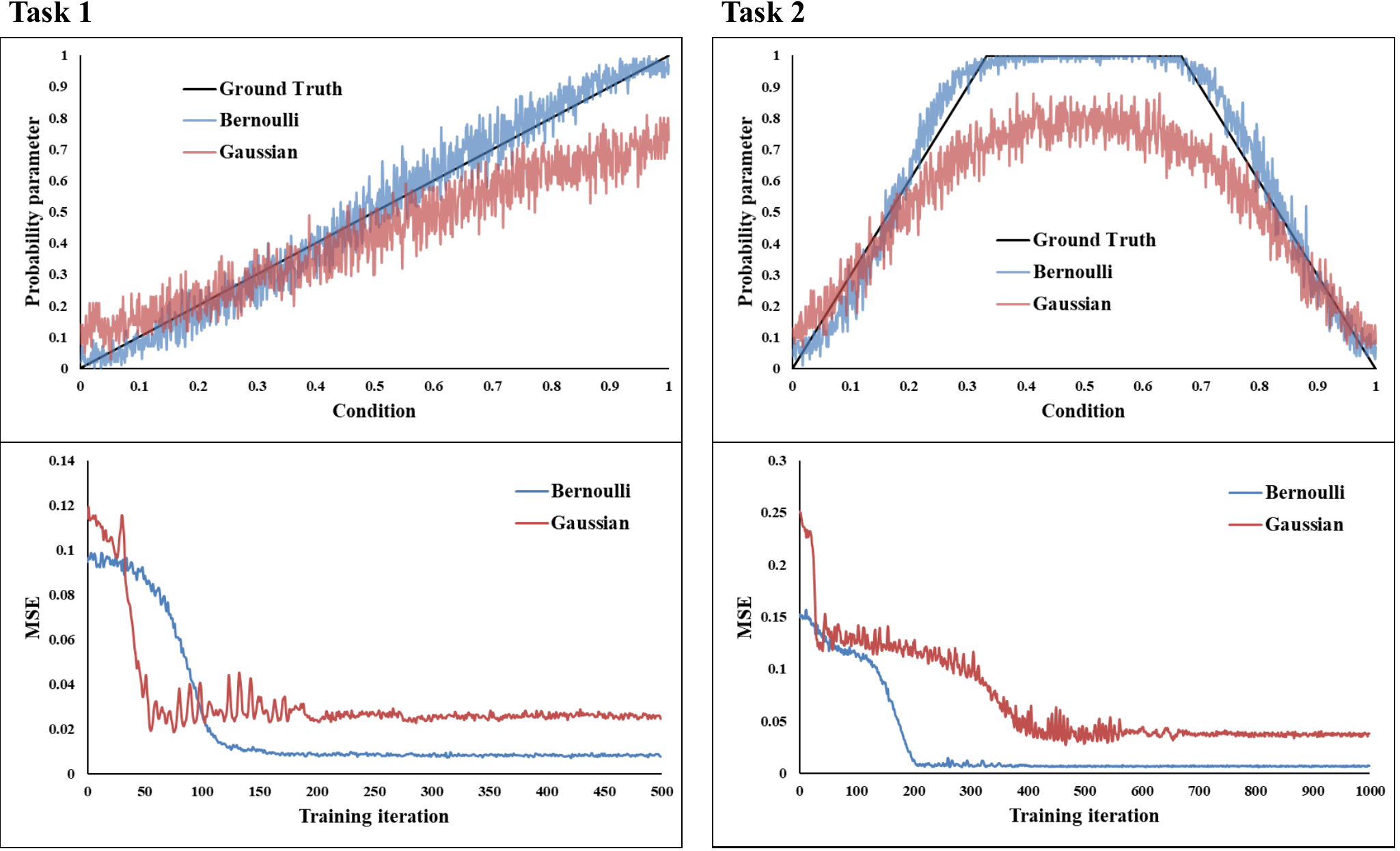}
  \caption{\textbf{1D binary classification tasks.} }
  \label{fig:appendix_visualization_toy_supp}
\end{figure}

\section{Performance curves}
\label{appendix:performance_curves}
Here we present a detailed performance comparison between Bernoulli- and Gaussian-based diffusion models over training iterations in \figref{fig:appendix_fig_p_i_curve_supp}. Results show that employing Bernoulli noise leads to faster convergence and higher performance in contrast to Gaussian noise.

\section{Toy example}
\label{appendix:toy_example}
We provide intuitive insight into why Bernoulli diffusion outperforms Gaussian diffusion by designing and conducting experiments on two simple one-dimensional binary classification tasks. To create the datasets, we apply a predefined conditional probability function, as shown in the first row of \figref{fig:appendix_visualization_toy_supp}, to map an input $x\in[0,1]$ to an output $y\in\{0,1\}$, which serves as the Ground Truth. For each data configuration, we use a 4-layer MLP for the noise estimation network. We train all runs using a learning rate of 1e-3. During inference, we uniformly sample 1000 points from $[0,1]$ and generate 100 samples using the diffusion model for each point. We take the mean of the 100 samples as the estimated probability parameter and evaluate performance using mean-squared error (MSE) between the estimated and ground truth probability parameters.  \figref{fig:appendix_visualization_toy_supp} shows that Bernoulli diffusion offers superior training stability, faster convergence, and better fitting of the conditional distribution compared to Gaussian diffusion.

\section{More examples of diverse segmentation masks}\label{appendix:more_example}
\begin{figure}[htbp]
\centering
  \includegraphics[width=0.9\linewidth]{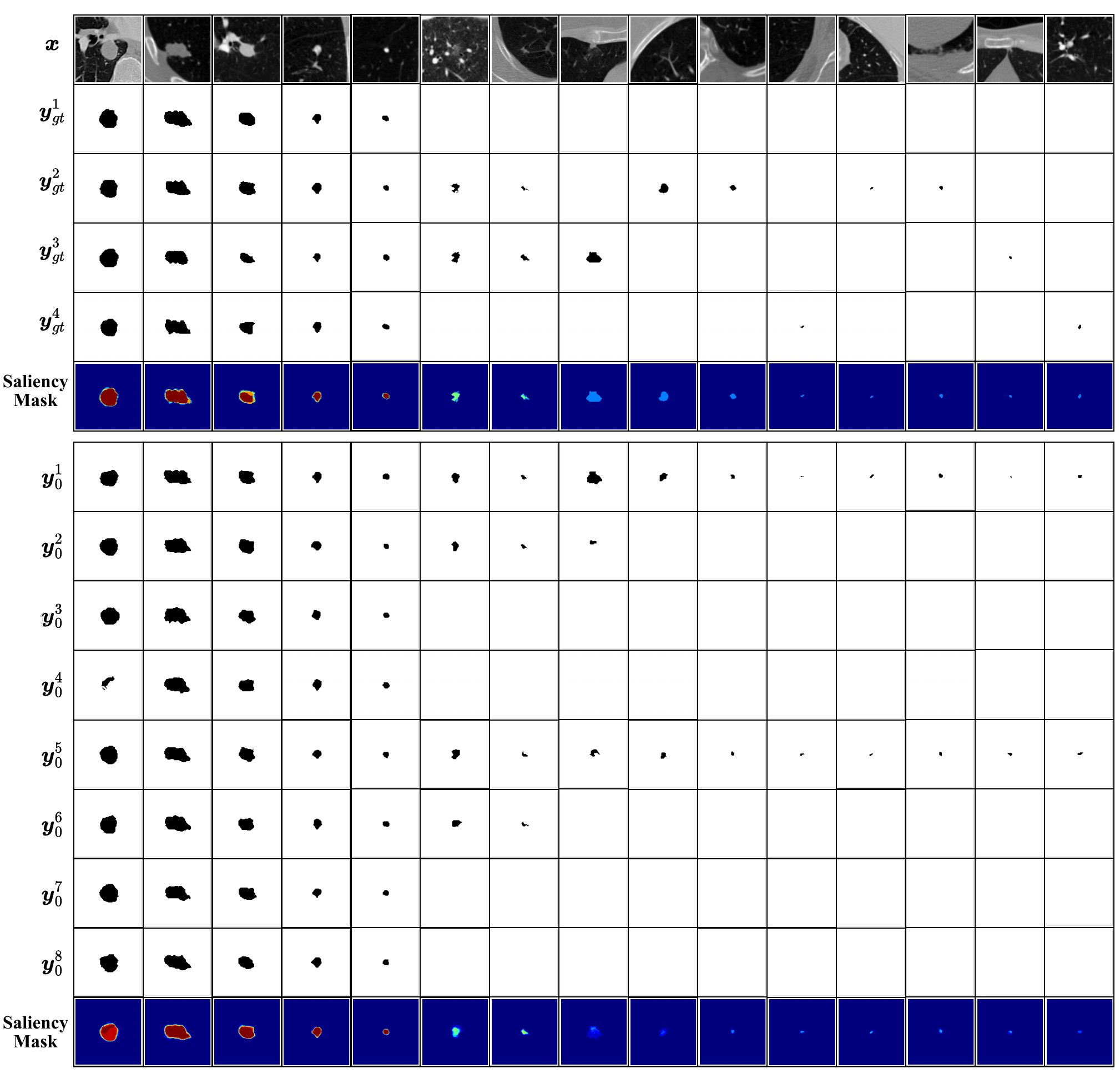}
  \caption{\textbf{More segmentation masks generated by our \methodname on LIDC-IDRI.} $\img{x}$ represents the input medical image. $\img{y}^{i}_{0}$ and $\img{y}^{i}_{\text{gt}}$ refer to the $i$-th generated and ground-truth segmentation masks, respectively. Saliency Mask is the mean of diverse segmentation masks.}
  \label{fig:appendix_visualization_LIDC_supp}
\end{figure}

\begin{figure}[htbp]
\centering
  \includegraphics[width=0.9\linewidth]{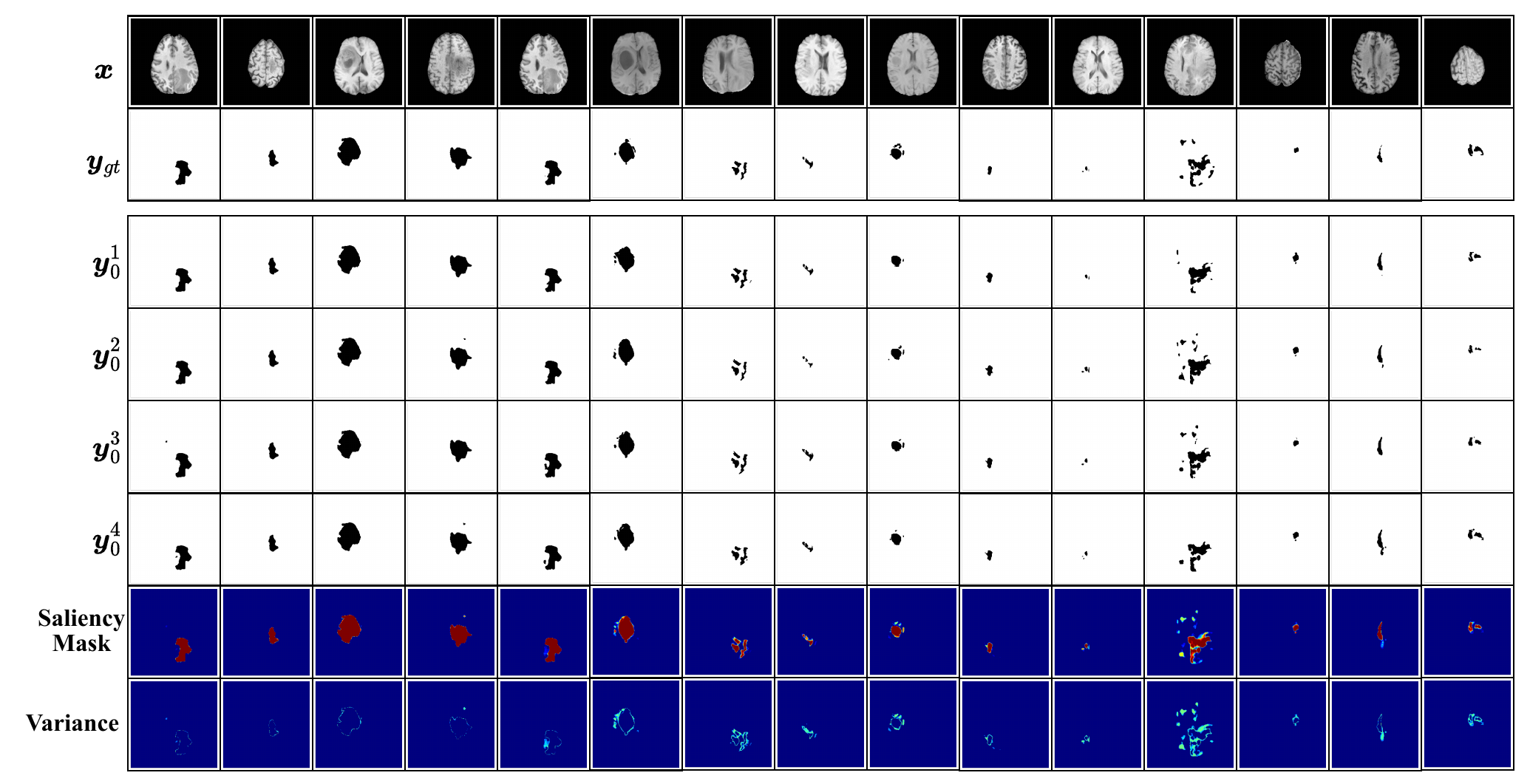}
  \caption{\textbf{More segmentation masks generated by our \methodname on BRATS~2021.} $\img{x}$ represents the input medical image, while $\img{y}_{gt}$ denotes the corresponding  ground-truth segmentation mask. $\img{y}^{i}_{\text{0}}$ refers to the $i$-th generated  segmentation mask. Saliency Mask is obtained by calculating the mean of diverse segmentation masks. Note that the variance of the generated segmentation masks is presented in the last row.}
  \label{fig:appendix_visualization_BRATS_supp}
\end{figure}

In this section, we provide additional examples of diverse segmentation masks generated by our \methodname for LIDC-IDRI and BRATS~2021, as shown in Figs.~\ref{fig:appendix_visualization_LIDC_supp} and~\ref{fig:appendix_visualization_BRATS_supp}, respectively.


\end{document}